**Epistemic NP Modifiers**

Dorit Abusch and Mats Rooth
*IMS, University of Stuttgart*

**1. Introduction**

Consider the sentence in (1a), where the adjectival passive *unknown* is a modifier within an existential NP. (1a) is ambiguous between a local reading (1b) and a propositional reading (1c).

(1)  a. Solange is staying in an <u>unknown</u> hotel.
   b. Solange is staying in a hotel. It is not a well-known one.
   c. Solange is staying in a hotel, and it is not known which hotel she is staying in.

We are interested in the propositional reading of the participle (p-participle). In the paraphrases in (1-4), what corresponds to the p-participle operates on a constituent question or a *that*-clause.

(2)  The suspects are in custody at two <u>undisclosed</u> locations.
  'The suspects are in custody at two locations and it is not disclosed which locations they are in custody in'.

(3)  The suspects were arrested at <u>unspecified</u> locations.
  'The suspects were arrested at some locations and it was unspecified which locations they were arrested in'.

(4)  Fabienne put the money in an <u>unexpected</u> place.
  'Fabienne put the money at some place, and it was not expected that she would put the money in that place.

 The readings under discussion are possible with singular indefinite descriptions (1 and 4), cardinal determiners (2), and bare plurals (3). In addition, with certain participles a similar reading is observed for definite descriptions:

(5)  Fabienne put the money in the <u>predicted</u> place.

However, in examples involving other quantificational determiners, the propositional reading is absent. Sentence (7) cannot mean that the campus police installed burglar alarms in most buildings and that it is unknown which buildings the campus police installed burglar alarms in.

(6)  Solange has stayed in every <u>unspecified</u> hotel.

(7)  The campus police installed burglar alarms in most <u>unknown</u> buildings.

This is suggestive of the distinction made in DRT theories between indefinite and definite descriptions on the one hand, and so called genuine quantifiers on the other.

## 2. The Straightforward Approach

In existing analyses of modification within NP, an adjective in the configuration [Det __ N] operates either just on a variable (as in the case of an extensional adjective) or on the intension of N (as in the case of the intensional adjectives *fake* and *former*). Thus if the clausal paraphrases in (1-4) are a guide to compositional structure, a novel compositional mechanism appears to be motivated. Let us assume the LF (8a) for (1), where the NP *an unknown hotel* has been assigned scope. It is possible to derive the desired interpretation for this representation by assuming that *unknown* is the main function, combining with an N', a DET, and the lambda abstract. If we want to arrive at the paraphrase in (1), the appropriate meaning for *unknown* is the one named by the lambda term (8b). This is the basis for the equivalence of logical forms (8c).

(8)
a.

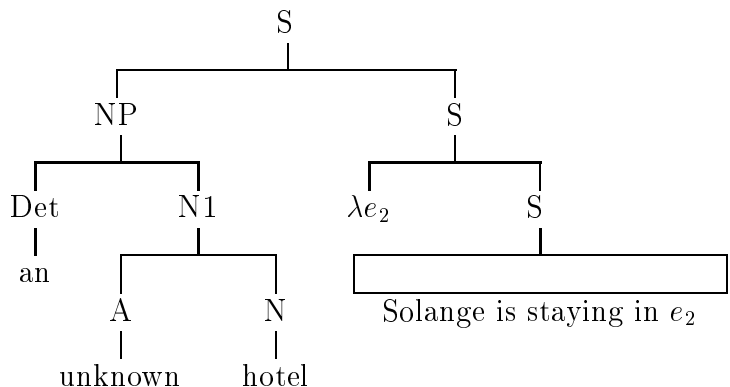

b. $\lambda Q \lambda D \lambda P[D(Q)(P) \wedge \mathbf{unknown}(\mathbf{wh}(Q)(P))]$
c. [[an unknown hotel] [$\lambda e_2$ [Solange is staying in $e_2$]]] $\equiv$
   [[a hotel] [$\lambda e_2$ [Solange is staying in $e_2$]]] and
   [it is unknown [[what hotel] [Solange is staying in $e_2$]]]

Making a modifier in a NP the main function is certainly a novelty, and it is clear that this analysis has a stipulatory character. What is worse, it does not explain the restriction on determiners. (8b) spells out the host clause and the indirect question as separate conjuncts, and it does so in a way which is independent of the specific determiner (corresponding to the variable D) which is involved. If we give (7) a logical form isomorphic to (8a), the analysis predicts the impossible reading described above.

## 3. Scope Ambiguities and Sensitivity to Attitudes

(9) can be understood either as: (i) it was not specified in the newspaper which city in Italy Solange has agreed to move to, or (ii) it was not specified in the agreement which city in Italy Solange would move to. In the first paraphrase, the embedding verb *agreed* is part of the indirect question and thus *unspecified* has wide scope over it. In the second paraphrase the indirect question involves only the clause *Solange would move to*. Thus we can say that *unspecified* has scope only over this clause, and characterize this as a narrow scope reading for *unspecified*.

(9)   There was a newspaper story about Solange. She has agreed to move to an unspecified city in Italy.

Curiously, when we change the participle the narrow scope reading is lost. (10) has only the wide scope reading: it was not disclosed in the newspaper which city in Italy Solange has agreed to move to.

(10)   There was a newspaper story about Solange. She has agreed to move to an undisclosed city in Italy.

With other participles we get only the narrow scope reading. The force of *undetermined* in (11) is: it is undetermined in the plan which University in China Salt 11 will be held at. The wide scope reading, namely that it is not specified in the announcement which University in China they are planning to hold Salt 11 at, is absent.

(11)   There was an announcement from the Salt steering committee. They are planning to hold the 11th Salt meeting at an undetermined University in China.

We have said that it is possible to think of these ambiguities as scope ambiguities. But they can also be thought of as ambiguities of an anaphoric character in the attitude picked up by the participle. In reading (ii) for (9), *unspecified* picks up the attitude of the agreement, and in reading (i), that of the newspaper report.

Instead of the scope of the participle relative to the embedding verb, one can consider the scope of the indefinite description relative to the embedding verb. In (10), a reading where the indefinite description headed by *city* has scope outside the complement of the embedding verb *agreed* is strongly favored. In (11), a reading where the indefinite description headed by *University* has scope inside the complement of the embedding verb *planning* is strongly favored. Thus the choice of a particular participle in an indefinite description can have the effect of disambiguating the scope of the indefinite description. (However, in section 7, we will see that the relation between the understood reading and the structural scope of the indefinite description is not necessarily the obvious one.)

### 4. Pronominal Paraphrases

We saw that sentences (1-4) have a clausal paraphrases (CP) where the participle operates on meaning elements contributed by all of its host clause. An alternative is a pronominal paraphrase (PP) which uses a pronoun to pick up a discourse referent whose identity is claimed to be unknown. This is illustrated below.

(12)   Solange is staying in an unknown hotel.
          CP: Solange is staying in a hotel, and it is not known which hotel she is staying in.
          PP: Solange is staying in a hotel x and it is not known which hotel x is.

Other examples of pronominal paraphrases are:

(13)   The suspects were arrested at unspecified locations.
          PP: The suspects were arrested at some locations and it was unspecified which locations they are.
(14)   Solange was at the party with an undisclosed man.
          PP: Solange was at the party with some man x and it is not disclosed which man x is.

In some cases, the pronominal paraphrase seems more correct than the clausal one. Suppose the label on the lime pickle says: limes, oil, red chillies, chemical preservative EDTA 1 mg, other chemical preservative 2 mg, nothing else. Solange reads the label and is not sure she wants any more. She says:

(15) a. This lime pickle contains an unspecified chemical preservative.
   b. CP: This lime pickle contains a chemical preservative and it is not specified which chemical preservative it contains.
   c. PP: This lime pickle contains a chemical preservative x and the label fails to specify which preservative x is.

While sentence (15a) is judged to be true in the scenario, speakers hesitate in their judgements of the clausal paraphrase (15b). Is it true (because of the 2 mg of unspecified chemical preservative which the lime pickle contains), or false (because of the 1 mg EDTA which the lime pickle contains)? The pronominal paraphrase (15c) seems to capture the meaning of (15a) unproblematically.

## 5. Discourse Referent Predication

Paraphrases involving pronouns suggest a different strategy for compositional interpretation, where the participle supplies a predication on a discourse referent, instead of having compositional scope over the host syntactic clause. To motivate the semantics, we will develop a connection between the sentences we want to analyze (such as 16a) and sentences such as (16b) where an epistemic modal *might* has scope over a dynamically bound pronoun.

(16) a. Solange is staying in an unknown hotel.
   b. Solange is staying in a hotel. It might be the Hotel Colbert. It might be the Days Inn.

Sentences of the second kind have been discussed and analyzed by Groenendijk, Stokhof, and Veltman (1996). In the dynamic quantified modal logic they present, (16b) is formalized as in (17).

(17) $\exists x_3[\text{hotel}(x_3) \ \& \ \text{stayin}(s,x_3)] \wedge \Diamond[x_3 = \text{colbert}] \wedge \Diamond[x_3 = \text{days}]$

The semantics for such formulas is stated in terms of the information states induced by successive conjuncts. An information state combines information about the world with information about variables; in this discussion, we will take an information state to be a set of world-assignment pairs, together with a specification of the set of available discourse referents, i.e. a file in the sense of Heim (1983).[1]

The file induced by the first conjunct in (17) consists of pairs $<g,w>$ such that in w, $g(x_3)$ is a hotel that Solange is staying in. The second conjunct $\Diamond[x_3=\text{colbert}]$ imposes a condition on this file: there are possibilities $<g,w>$ where $g(x_3)$ is the Hotel Colbert. This modal condition functions as what is known as a test: if the condition is satisfied, the output file for the middle conjunct is the same as the input file, while if the condition is not satisfied, the output file is the absurd information state, reflecting contradictory information.

As a first cut at stating the semantics of the participles we are concerned with, consider (18b). The variable corresponding to *hotel* is $x_3$, and that $x_3$ is unidentified is expressed by the second conjunct: for any y, there is a possibility <g,w> where $g(x_3)$ is not y.

(18)  a. Solange is staying in an unidentified hotel.
      b. $\exists x_3[\text{hotel}(x_3) \wedge \text{stayin}(s,x_3)] \wedge \forall y \Diamond [y \neq x_3]$

Here is a corresponding formulation for identification rather than non-identification:

(19) a. (According to the report) Solange is staying in a hotel, which is identified.
     b. $\exists x_3[\text{hotel}(x_3) \wedge \text{stayin}(s,x_3)] \wedge \exists y \Box [y = x_3]$

Example (19a) includes reference to a report, the content of which is being described. (18a) should in fact be thought of in a similar way, as conveying part of the content of some report, attitude, or other informational entity. Suppose that a memorandum $M_1$ includes the sentence *Solange is staying in the Days Inn*. Then sentence (18a) is intuitively false as a description of the content of $M_1$, while sentence (19a) is intuitively true, because the report does identify the hotel that Solange is staying in. Suppose memorandum $M_2$ includes the sentence *Solange is staying in a hotel*, and no further information about the hotel. Sentence (18a) is true as a description of the content of $M_2$, while (19a) is false as a description of the content of this memo.

In modeling these intuitions in the semantics, we follow Groenendijk, Stokhof, and Veltman (1996) in assuming that, in the construction of models, the same individuals are used in stating the extension of predicates in different worlds, in a way which is significant for the semantics. For instance, we assume that the Hotel Colbert is a certain fixed element of the universe of individuals, in any world where the Hotel Colbert exists.[2]

With this assumption, the hotel variable $x_3$ being identified in an information state p can be analyzed as $g(x_3)$ having the same value in all possibilities <g,w> in p. For instance, if in each possibility <g,w> in p $g(x_3)$ is the Hotel Colbert, then the variable (or discourse referent) $x_3$ counts as identified in p. In GSV's logic, the variable $x_3$ being identified in the contextual information state can be formalized as follows:

(20)   $Id(x) =_{def} \exists y \Box [y = x]$

This leads to the following formalization of (19a):

(21)   $\exists x_3[hotel(x_3) \wedge stayin(s,x_3)] \wedge Id(x_3)$

In (21b), the Id formula in (21a) is rewritten using (20). A similar formalization can be given for a variable being unidentified:

(22)   a. $Ud(x) =_{def} \forall y \Diamond [y \neq x]$
       b. $\exists x_3[hotel(x_3) \wedge stayin(s,x_3)] \wedge Ud(x_3)$

(22a) defines an operator Ud which expresses the non-identification of the argument dref; this results in the formalization (22b) for (18a).

In these formalizations, the file in which the dref is required to be identified or unidentified is the contextual file provided by the dynamic semantics. Rather than using this dynamic setup with implicit information states, we prefer to work with a representation where the file is made explicit as a term in the semantic representation. From now on, we use p, q, $p_1$ and so forth as variables of the file type, and write $Id(x,p)$ for the dref x being identified in the file p, and $Ud(x,p)$ for x being unidentified in p. The semantics is as explained above: a dref x is (at least partially) unidentified in p iff there are possibilities <g,w> and <g',w'> in p where g(x) and g'(x) are different. The dref x is identified in p iff for all <g,w> and <g',w'> in p, g(x) is equal to g'(x).[3]

A second idea which we will use is a semantics for attitudes involving anaphoric reference to files. Consider the following sequence of two belief attributions.

(23)   Monique believes Solange has a lover$_7$.
       She thinks he$_7$ is a musician.

The second sentence picks up a discourse referent introduced by the narrow-scope indefinite description *a lover* in the first one. This is an attitudinal version of the phenomenon of modal subordination (Roberts 1987, 1996). Following Guerts (1995), we will work with an analysis of (23) where *think* in the second sentence has as an implicit argument a file introduced by the complement in the first sentence. The notation below is that of Geurts, which consists of a linear DRT representation combined with the file increment notation of Heim (1982, 1983).

(24)  $p_2:p_1 + [x_7|lover(x_7,s)]$   $p_3:p_2 + [\ |musician(x_7)]$
      believe(m,$p_1$)   believe(m,$p_2$)   believe(m,$p_3$)

On the first line, $p_2$ is defined as the result of updating a contextual file $p_1$ with the information that Solange has a lover; similarly, $p_3$ is the result of updating $p_2$ with the information that he (i.e. $x_7$) is a musician. On the second line, all these three files are described as being believed by Monique, by means of a relation *believe* between an individual and a file. A representation along these lines gives a straightforward account of how an antecedent for the pronoun in (23) is set up: this follows from the fact that the input file $p_2$ for the file change potential [ | musician($x_7$)] has a value for the dref $x_7$.

To state the lexical semantics of *believe* we assume: (i) x's beliefs are characterized by the set of worlds consistent with her beliefs, which we write belief(x). (ii) x believes a file p iff the world projection of p is exactly belief(x). We write this relation belief(x) $\approx$ p, and where it holds, we say that p is a world-preserving update of belief(x).

The lexical semantics for *believe* is stated in terms of a presupposition and an assertion:

(25)  x believe$_p$ $\phi$    presupposes   belief(m) $\approx$ p
                             asserts        belief(m) $\approx$ p + $\phi$

Here x is the individual contributed by the subject NP, $\phi$ is the file change potential contributed by the complement, and p is an implicit file argument. The presupposition constrains p to be some world-preserving update of belief(m). The assertion requires that updating p with $\phi$ eliminates no worlds. Restating this in the notation introduced above, and binding the arguments with lambda, we obtain the following lexical semantics for *believe*:

(26)  $\lambda\phi\lambda x[p'|\partial belief(x) \approx p,\ p':p + \phi,\ belief(x) \approx p']$

The presupposition is rendered using the partial operator (Beaver 1995); the contextual file is p, and the new file introduced is p'. That p' is novel is expressed by putting it into the domain of the box term, before the vertical line.

Our aim now is to analyze the following example along similar lines; notice that the second sentence is understood as conveying the content of the story, in a kind of free indirect discourse.

(27)  There was a story$_4$ in Variety.
      Solange has signed with a Hollywood agent, who is not identified.

The first sentence introduces an individual-level discourse referent $x_4$ for the story. We assume that informational entities such as stories have propositional contents, and that these are introduced into the discourse representation using a function mapping the individual story to its content: the propositional content of $x_4$ is content($x_4$). In the representation below, the description of the content of the story is handled in the same way as belief attributions were handled above.[4]

(28) $[p_2|\ p_2{:}\text{content}(x_4)+[x_5|\ \text{hollywood-agent}(x_5),\ \text{sign-with}(s,x_5)],$
$\text{content}(x_4){\approx}p_2]$

The contribution of the non-restrictive relative clause *who is not identified* can now be captured with the formula $Ud(x_5,p_2)$ asserting that the discourse referent $x_5$ is not identified in the output file $p_2$. As we explained earlier, this means that the value of the assignment on $x_5$ is different in different possibilities in $p_2$. If the story had in fact said that Solange had signed with a Hollywood agent named Sheldon Plotkin, then in each possibility $<g,w>$ in $p_2$, $g(x_2)$ would be the same individual, assuming a background context in which naming is sufficient to identify individuals. In this case, the relative clause *who is not identified* is predicted to be false, matching our intuitions.

Summing up the analysis, predications of identifiedness and unidentifiedness are analyzed as relations between a discourse referent and the output file introduced by the description of content. This predication is a separate conjunct with the content description. The complete representation for the second sentence in (27) is:

(29) $[p_2|\ p_2{:}\text{content}(x_4)+[x_5|\ \text{hollywood-agent}(x_5),\ \text{sign-with}(s,x_5)],$
$\text{content}(x_4){\approx}p_2,\ Ud(x_5,p_2)]$

In sentence (27), the semantics of unidentifiedness is introduced by a predicate in a non-restrictive relative clause. The examples we discussed earlier, however, involved participles in modifying positions in the NP. Our strategy is to assimilate the modifying structure (30) to the non-restrictive one.

(30) She$_1$ has signed with an unidentified Hollywood agent$_5$.

Let us assume that *unidentified* in (30) denotes the property $\lambda v Ud(v,p)$, where for the moment we take the file argument to be a free variable. In order for this to have the right meaning, it is essential that $v$ be a variable with the type of discourse referents, rather than the type of individuals. It is discourse referents which are matched across possibilities in files, and thus have the potential of being identified or unidentified across possibilities. However, it is not obvious that a variable for discourse referents should be available at the level of the participle, assuming a syntactic structure [an [unidentified [Hollywood agent]]]; one might suppose that the discourse referent is introduced by the article, and that the argument of the noun *Hollywood agent* is simply an individual. This is the choice made in Rooth (1987), for instance.  However, certain compositional DRT theories such as Asher (1993) and Muskens (1996) propose type-raised denotations for nouns which take discourse referents rather than individuals as arguments.   In Muskens' system, a raised noun denotation has the type [π], characterizing a one-place property of discourse referents; π is the type of discourse referents.  According to Muskens (p.c.), this choice was dictated by considerations of elegance, rather than any concrete empirical argument.  If the analysis now being discussed is correct, p-participles  provide empirical support for this compositional approach.

The remaining thing to pin down in giving an analysis of (30) is how the participle *unidentified*, once it has picked up its discourse referent argument, is integrated into the semantic representation of the host sentence.  One possibility is that modifiers can be lexically specified as having a non-restrictive compositional semantics, meaning that they combine as conjuncts with the semantic representation of the host sentence.  This would allow us to mimic the analysis of the example with a non-restrictive relative clause.  Something has to be said about the syntax-semantics interface for non-restrictive modifiers: one option is a storage,  where  the meaning of a non-restrictive element is stored and  conjoined at a higher level.   If we make this move, the file argument p of *unidentified*  can be treated as a free variable, which picks up its antecedent anaphorically.  This is the solution which we tentatively adopt,  largely on the grounds of simplicity.

The other possibility is that the p-participle is tied in more tightly with the semantics of the host clause. In particular, perhaps the file argument is picked up compositionally rather than anaphorically.[5]  We will not pursue this idea here, although we have no reason to dismiss it.

**6. Sensitivity to Attitudes**

In section 3 we pointed out that choosing of a particular p-participle can have a scope disambiguating effect.   In the example below, (31a) is a context sentence and (31b,c,d) are alternative continuations.   The modifier *undisclosed* disambiguates the scope of the indefinite description *a team in Italy* in the direction of scope superior to *agreed*, while the modifier *undetermined* disambiguates in the direction of scope inferior to *agreed*, i.e. gives minimal scope for the indefinite description. Finally, *unspecified* is compatible with either scope.

(31) a. There was a press release about Solange, saying that she will be leaving Manchester.
 b. She has made an agreement to play for an undisclosed team in Italy.
 c. She has made an agreement to play for an undetermined team in Italy.
 d. She has made an agreement to play for an unspecified team in Italy.

We suggested that scope disambiguation is to be understood as a matter of compatibility between attitudes and participles. In the case of the attitudes and participles involved in (31), we conjecture that the following table of compatibilities underlies the scope data.

(32)   undisclosed      press release   *agreement
       undetermined    *press release    agreement
       unspecified      press release    agreement

Presumably these compatibilities and incompatibilities derive from the lexical semantics of the base verb. For instance, a communication such as a press release may disclose which team Solange will play for, but because a communication does not involve the requisite element of planning and commitment, it cannot determine which team she will play for.[6]

To capture sensitivity to attitudes, we add an additional contextual argument to the participle, corresponding to the individual-level variable for the story or agreement, which is written as r in the descriptions of the lexical semantics below.

(33) a. undisclosed(v,r,p)    presupposes   r is a communication
                                            content(r) $\approx$ p
                              asserts       Ud(v,p)
     b. undetermined(v,r,p)   presupposes   r is a plan
                                            content(r) $\approx$ p
                              asserts       Ud(v,p)
     c. unspecified(v,r,p)    presupposes   content(r) $\approx$ p
                              asserts       Ud(v,p)

The attitude restriction is stated as a presupposition regarding the informational individual r.

We assume that (31b) has two logical forms, differing in the scope of the indefinite description headed by *team*. Leaving out the p-participle, these are the wide-scope LF (34), and the narrow scope LF (35).[7]

(34) [[a team in Italy]$_4$ [Solange has made [an agreement [to play for e$_4$]]]]

(35) [Solange has made [an agreement [[a team in Italy]$_4$ [to play for e$_4$]]]]

Assuming a context where (34) is understood as conveying the content of a press release r, we obtain the following representation on the theory outlined above.

(36) [p$_1$|  content(r) ≈ p$_1$

　　　　p$_1$:content(r) + [p$_2$, x$_1$, x$_2$|   team-in-italy(x$_1$)
　　　　　　　　　　　　　　　　　　　　　　agreement(x$_2$)
　　　　　　　　　　　　　　　　　　　　　　make(solange,x$_2$)
　　　　　　　　　　　　　　　　　　　　　　content(x$_2$) ≈ p$_2$
　　　　　　　　　　　　　　　　　　　　　　p$_2$:content(x$_2$) + [|play-for(solange,x$_2$)]]

　　　　]

The wide scope for *a team in Italy* is reflected in the fact that the discourse referent x$_1$ is quantified outside the file change potential which updates the content of the agreement x$_2$.

The participle *undisclosed* in (31b) generates the formula (37a), which is stored and retrieved at some level or other.

(37)　a. ∂communication(y), ∂content(y) ≈ q, Ud(x$_1$,q)
　　　b. ∂communication(r), ∂content(x$_1$) ≈ p$_1$ , Ud(x$_1$,p$_1$)

Here y and q are discourse referents which must find antecedents. Generating from (36) a representation which includes the contribution of the participle involves:

 (i) Choosing a scope for the stored formula.
 (ii) Picking an antecedent for the individual discourse referent y.
 (iii) Picking an antecedent for the file discourse referent q.

These choices are subject to various constraints. First, the presuppositions in (37a) constrain the antecedents for y and q. Second, an antecedent for q must be chosen which has the dref x$_1$ in its domain; this can be regarded as a presupposition of the formula Ud(x$_1$, q). And presumably, as a result of storage, the stored formula must end up either at the level of *team-in-italy(x$_1$)* in (36), or at some superior level. These conditions are satisfied if the file q is identified with p$_1$, the informational individual y is identified with r, and the landing site for the resulting formula (37b) is the position marked with the little feet in (36). Consider the various presuppositions:

(i) ∂communication(r) is satisfied because we are assuming that r is a press report, and that press reports are communications.

(ii) ∂content(r) ≈ $p_1$ is satisfied because it occurs in a conjoined block with the asserted formula content(r) ≈ $p_1$.

(iii) $p_1$ is the output file for an update which introduces the dref $x_1$. Therefore the constraint that $x_1$ be in the domain of $p_1$ is satisfied.

In summary, (36) with the indicated scope for (37b) is a grammatically possible representation, which is compatible with the presuppositions introduced by the participle. In other words, the participle *undisclosed* is compatible with wide scope for *a team in Italy*. Note that the fact that the presuppositional constraints are satisfied does not mean that the participle adds no information about the press report; to the contrary, it adds the information that the press report does not identify the team in Italy that Solange has agreed to play for.

Consider the effect of substituting *undetermined* for *undisclosed*, keeping everything else the same. In this case, the formula contributed by the participle is:

(38)  ∂plan(r), ∂content(r) ≈ $p_1$, Ud($x_1$,$p_1$)

We assume that the constraint ∂plan(r) is incompatible with the fact that r is a press report. The result is that the representation is semantically filtered. So we have derived the fact that (31c) lacks the reading with wide scope for *a team in Italy*.

Symmetrically, consider the semantic representation reflecting narrow scope for the indefinite description:

(39)  [$p_1$| content(r) ≈ $p_1$

$p_1$:content(r) + [$p_2$, $x_2$|    agreement($x_2$)
                                    make(solange,$x_2$)
                                    content($x_2$) ≈ $p_2$
                                    $p_2$:content($x_2$) + [$x_1$|  play-for(solange,$x_2$)
                                                                    team-in-italy($x_1$) ]
                                            ]]

We start by considering a version with *undetermined*, a file argument $p_2$, and an individual argument $x_2$. This results in the following as the formula contributed by the participle.

(40)   $\partial\text{plan}(x_2), \partial\text{content}(x_2) \approx p_2, \text{Ud}(x_1,p_2)$

If the landing site for the stored formula (40) is the position marked by the feet in (39), the presuppositions are satisfied because (i) the asserted formula content($x_2$) ≈ $p_2$ is conjoined with $\partial$content($x_2$) ≈ $p_2$; (ii) agreements, we stipulate, are plans; (iii) because $p_2$ is the output file for an update introducing the dref $x_1$, $x_1$ is in the domain of $p_2$. Thus *undetermined* is compatible with narrow scope for *a team in Italy*. Notice that the information it adds involves an embedded attitude: according to the report, the agreement does not determine which team Solange will play for.

If we substitute *undisclosed* for *undetermined* in the narrow-scope grammatical analysis, we derive a presupposition $\partial$communication($x_2$). Because agreements are not communications, the representation is filtered. So we have derived the fact that (31b) is incompatible with narrow scope for the indefinite description. (31d) is compatible with either scope, because *unspecified* introduces no presupposition which is incompatible with either a press report or an agreement attitude.

To complete the argument, one has to show that no other landing sites for the participles are possible. In the representations above, there are indeed no other possibilities, because of the requirement that $x_1$ be in the domain of the file argument of the participle.

## 7. The Scope of Specifics

The example below involves a positive participle, and is understood to entail that the agreement specifies a particular team that Solange is supposed to play for.

(41)   Solange has made an agreement to play for a specified team in Italy.

The understood semantics might lead us to think that this is a reading where the indefinite description headed by *team* has scope over *made an agreement*. But our analysis tells us that the indefinite description can have minimal scope. The representation is a variation on the narrow scope representation from the previous section:

(42) $[p_1|\ \text{content}(r) \approx p_1$

$p_1:\text{content}(r) + [p_2, x_2|\quad \text{agreement}(x_2)$
$\text{make}(\text{solange}, x_2)$
$\text{content}(x_2) \approx p_2$
$p_2:\text{content}(x_2) + [x_1|\ \text{play-for}(\text{solange}, x_1)$
$\text{team-in-italy}(x_1)\ ]$
$\partial \text{content}(x_2) \approx p_2$
$\text{Id}(x_1, p_2)\quad ]]$

The last two formulas represent the contribution of *specified*, assuming appropriate antecedents for the free variables. The formula $\text{Id}(x_1,p_2)$ says that the dref $x_1$ is identified in the file $p_2$; the force of this is that the agreement contains enough information to specify the team uniquely.

It turns out that a representation with wide scope for the indefinite description is also possible:

(43) $[p_1|\text{content}(r) \approx p_1$

$p_1:\text{content}(r) + [p_2, x_1, x_2|\quad \text{team-in-italy}(x_1)$
$\text{agreement}(x_2)$
$\text{make}(\text{solange}, x_2)$
$\text{content}(x_2) \approx p_2$
$p_2:\text{content}(x_2) + [\ |\text{play-for}(\text{solange}, x_1)]$
$\partial \text{content}(x_2) \approx p_2$
$\text{Id}(x_1, p_2)\quad ]]$

The formulas $\partial \text{content}(x_2) \approx p_2$ and $\text{Id}(x_1,p_2)$ are in the same place as it is in (42); this is possible because the participle meaning is stored and may be retrieved at any level.[8]

Are the narrow scope specifics predicted by the current theory a reality? Below, we apply some independent diagnostics.

(44) Spy has made the claim that Solange is involved in a triangular affair with an unidentified man and an identified female relative of his.

(45) Spy has made the claim that there is an Andean guerrilla leader, who is not identified, hiding out in Solange's place in Brittany.

(46) Spy has made the claim that there is an unidentified Andean guerrilla leader hiding out in Solange's place in Brittany.

In (44), the idea is that *an unidentified man* caps the scope of *an identified female relative of his*, because the latter contains a pronoun anaphoric to the former. On the assumption that the former must have narrow scope (we expect this is so, though we do not take it for granted), the latter must have narrow scope also. Note that it is perfectly possible for the Spy story to not identify the man, but to identify the relative by naming her. (45) involves there-insertion, which Heim (1987) shows entails narrow scope in LF. This example, with the participle in a non-restrictive relative clause, is impeccable. We are not sure whether the corresponding modifier example (46) is good, on the relevant reading where *unidentified* comments on the content of the claim, rather than being part of the claim.

**8. Determiners**

The examples below illustrate the constraint on determiners which we discussed in section 1. The hash marks in (49) indicate that a propositional reading for the participle is absent.

(47)   There was a story in the Times about Solange.
(48)   a. She has paid a bribe to <u>an</u> unspecified member of Congress.
       b. She has paid bribes to <u>two</u> unspecified members of Congress.
(49)   a. #She has paid bribes to <u>most</u> unspecified members of Congress.
       b. #She has paid bribes to <u>almost every</u> unspecified member of Congress.

As we said, this calls to mind the semantic distinction made in DRT theories between indefinite descriptions and genuine quantifiers. We initially thought that this would provide an immediate explanation for the constraint on determiners in the analysis now being considered. For the formula Ud(x,p) to be meaningful, x must be in the domain of the file p. And it is a classic tenet of DRT theories that indefinite descriptions, but not genuine quantifiers, contribute discourse referents to their output context.

   Things are not quite this simple, though, because genuine quantifiers do set up group-level discourse referents:

(50)   Solange paid bribes to most members of Congress.
       In return, they supported tax relief for foreign movies.

And in fact, discourse referents introduced in this way can be arguments of epistemic participles (or the corresponding verbs---the underlined words below could be either verbal or adjectival participles).

(51) According to the story, Solange has paid bribes to most members of Congress. They are <u>named</u>/<u>identified</u> in the story.

This example contrasts minimally with (52) where the participle is in a prenominal modifying position, rather than predicate position in a separate sentence. (52) does not have a propositional reading, though there is a perfectly coherent thing for it to mean, namely what (51) means.

(52) Solange has paid bribes to most #<u>named</u>/#<u>identified</u> members of Congress.

The problem with (52) must somehow have to do with compositional semantics. In the analysis we outlined, the compositional environment of *identified* in (52) is roughly as follows.

(53)  [most [λx[identified(x,r,p)]  members of congress]]

The participle picks up a discourse referent argument compositionally, resulting in the following stored formula:

(54)  ∂a(r), ∂content(r) ≈ p, Id(z,q)

Here ∂a(r) expresses the constraint regarding the attitudes compatible with *identified*, whatever this is exactly, and z is the dref of which identifiedness is predicated. The problem with (52), we suggest, is that the group discourse referent (corresponding to the maximal group of congressmen to whom Solange has paid bribes) is not available at this compositional level. In particular, this group discourse referent is not the compositional argument of *members* which is picked up by *identified*. The effect is that a formula equivalent with the second sentence of (51) is not derived compositionally. This comports with the analysis of Kamp and Reyle (1993), where the group discourse referent is derived at the DRS level by a structure-building operation called summation, rather than directly in the compositional semantics of the quantified NP.

The above is just a sketch of an analysis, which has to be confirmed by a detailed statement of the semantics; we leave this as a promissory note. But if the analysis (or something along the same general lines) is correct, it is extremely intriguing, because it introduces issues of sub-clausal compositional semantics into the current debate about typed dynamic systems for natural language interpretation.

## 9. Pronominal Subjects and Lexical Structure

In the previous section, we saw in (51) an example of a p-participle occurring with a pronominal subject. Such examples are possible with some specific lexical items, but not others:[9]

(55) a. According to Spy Magazine, Solange was at the Mayor's party with a bald muscular man. He {?? is unidentified/ is not identified}
b. According to Spy Magazine, Solange was at the Mayor's party with an unidentified muscular man.

(56) a. (According to the story) Solange was at the ball with a British nobleman. He {? is unnamed /is not named}.
b. Solange was at the ball with an unnamed British nobleman.

(57) a. According to the proposal the fragment will be written by two graduate students. They {??are unlisted/ are not listed}.
b. ? The fragment will be written by two unlisted graduate students.

(58) a. According to the plan phonetic data will be collected on five Bantu languages. They {??are unenumerated/ are not enumerated}.
b. Phonetic data will be collected on five enumerated Bantu languages.

(59) a. Solange was sighted at the ball with a British nobleman. He is {*unknown/ *not known}. (only local reading).
b. Solange was sighted at the ball with an unknown man.

(60) a. According to the proposal the grammar fragment will be written by a graduate student. He {??is unspecified/ is not specified}.
b. The grammar fragment will be written by an unspecified graduate student.

(61) a. * Solange agreed to move to a town in Italy. It {* is undisclosed/ * is not disclosed}.
b. Solange agreed to move to an undisclosed town in Italy.

Although the data are not always clear, some of the examples with pronominal subjects, such as the second alternative in (55a), are definitely good while others such as (61a) are definitely bad.

In (55a,56a,57a,58a) there is a further distinction between versions with a morphological negation *un-,* and a syntactic negation *not.* In (55a), it seems to us that the version with *unidentified* does not quite succeed in commenting on the content of the story. In the variants below, we feel that both the present and past tense versions with syntactic negation (63a) and (63b) convey the relevant reading, while the present and past tense versions (63c,d) do not have the relevant reading, where they comment on the content of the story. (63e) is good, but the unidentifiedness here does not refer specifically to the content of the story.

(62) In July, there was a story in Spy saying that Solange was at the Mayor's party with a bald muscular man.

(63) a. He was not identified.
 b. He is not identified.
 c. He was unidentified.
 d. He is unidentified.
 e. He is still unidentified.

An approach to these interpretive contrasts was suggested in a question after our talk: the examples with pronominal subjects (which do have a reading conveying the content of a contextual attitude such as the Spy story) are verbal passives, rather than adjectival ones. This is not refuted by the fact that a present tense copula is possible in (63b), for this is also true of the transitive verbal example (64b).

(64) a. The story did not identify him.
 b. The story does not identify him.

Further, the contrast between acceptable and unacceptable pronominal arguments is replicated with transitive forms:

(65) a. *The story does not disclose him.
 b. The story does not name him.
 c. The proposal does not enumerate them.

Summing up these observations, it the propositional reading is not observed for adjectival participles in predicate position with pronominal subjects. Moreover, only some modifying p-participles have verbal counterparts with pronominal arguments. We will sketch an approach to these data involving lexical decomposition of p-participles. The analysis is motivated by a paraphrase. While (61a) and (65a) are bad, the desired reading can be expressed with a full noun phrase in the place of the pronoun:

(66) a. His identity is not disclosed.
 b. The story does not disclose his identity.

These are just a special case of definite descriptions with concealed question readings (e.g. Grimshaw (1977)):

(67)  a. His age is not disclosed.
     b. The story does not disclose his profession.

This suggests that the basic semantics for *disclose* does not convey the identificational meaning. Rather the morphological analysis of the p-participle reading of *undisclosed* is IDENTITY-un-disclosed or un-IDENTITY-disclosed, where a root expressing the identificational component of meaning is incorporated. Actually, one might want to go a bit further, and analyze concealed questions as incorporating an operator:

(68) The story does not CQ-disclose his profession.

The propositional reading of *undisclosed* should then incorporate CQ as well as IDENTITY. For the present, we will stipulate that the modifying p-participle is a distinct lexical item which incorporates the indentificational reading. An interesting project would then be to try to explain in morphological terms the distributional constraints on propositional readings which were noted above, in line with the program of Pesetsky (1995) regarding interactions between syntactic distribution, zero affixation, and lexical semantics.

A further point has to do with the distinction between verbs such as *identify* and verbs such as *disclose*. We suggest that verbs such as *identify* are denominal; *identify* has the analysis disclose-CQ-the-identity-of, and *name* has the analysis *disclose-CQ-the-name-of*. In this sense, they are comparable to *shell*, with the meaning of remove-the-shell-of. A denominal analysis should be the basis for deriving the goodness of the second alternative in (55a): the pronoun is an immediate argument of the overt root. Turning to the modifying participles *undisclosed* and *unidentified*, these are a lexical-structural doublet, modulo differences in the attitude conveyed: both are based on the compositional structure not-disclose-the-identity-of. In one case, the root *disclose* is overt, and in the other case the root *ident* is overt.

We will defer an investigation of these ideas to a later occasion.[10]

## 10. A Certain

Lauri Carlson (p.c.) has suggested that the semantics of *a certain* is to be explicated in terms of identificational questions. In (70), the contribution of *a certain* would be that the speaker can answer the question *which city is it?* This leads to paraphrases as in (71).

(70)   Solange has moved to a certain city in Italy.

(71)   a. Solange has moved to [a city in Italy]₂, and I could tell you which city it₂ is.

   b. Solange has moved to [a city in Italy]₂, and I know which city it₂ is.

It seems that *a certain* is an idiom, because it isn't possible to make substitutions in the position of the determiner:

(72)   a. *Solange has boyfriends in two certain cities in Italy.

   b. *Solange has boyfriends in several certain cities in Italy.

Nevertheless, the glosses (71) have the form of pronominal/indirect question paraphrases for p-participles. We will treat *a certain* as a single word, a determiner. As far as we can see, there is no reason to treat *a certain* as two words, although (unlike *another*) it is conventionally written as two words.[11] The determiner *acertain* has a meaning equivalent to the composition of the determiner *a* with some suitable p-participle. What attitudes is *a certain* compatible with? The example below shows that, like *specified*, it is compatible with a plan attitude.

(73)   Solange has made an agreement with the team.
   She will play for a certain team in Italy.
   I have no idea which team it is.

The third sentence shows that here *a certain* can pick up the attitude of the agreement, rather than the attitude of the speaker's beliefs. In the example below, we have a communication attitude.

(74)   There was a story in Spy about Solange.
   According to the story, she has moved to a certain remote island in the Pacific.
   I don't know which one, it was some exotic-sounding place.

The next examples show that *a certain* can pick up a belief attitude.

(75)   Claude evidently believes that Solange is involved with a certain ballet dancer.
   I have no way of telling who this dancer is supposed to be.

(76)   #Claude evidently believes that Solange is involved with a specified ballet dancer.

(76) shows that *specified* is not compatible with a belief attitude: it cannot be understood as meaning that the dancer is identified in Claude's beliefs. So, *a certain* seems to be less restrictive than *specified*.

The compatibility of *a certain* with a belief attitude is a basis for explaining what is going on in examples such as (70), where *a certain* is understood as relating to the speaker's beliefs. The reasoning would be that utterances are pragmatically interpreted as conveying the speaker's beliefs (or purported beliefs), and therefore this attitude is available to be picked up by *a certain*. This kind of use is the most common one, and is what is usually involved in the much-discussed scope disambiguating effects of *a certain*. In the example below, if *a certain* is to pick up the speaker's attitude, then the indefinite description headed by *member* has to have maximal scope.

(77) Nobody likes a certain member of the cast.

**11. Conclusion**

The main goal of our analysis was to account in an explanatory way for the constraint on determiners and for the scope-disambiguating effects of specific participles. Scope disambiguation was accounted for lexically: the lexical meaning of a p-participle encodes presuppositions which turn out to have a filtering effect on possible logical forms. To account for the constraint on determiners, we said that p-participles pick up a discourse referent argument compositionally. Combined with a standard idea from DRT/dynamic theories regarding a difference between indefinite descriptions and genuine quantifiers, and an assumption about the treatment of discourse referents in the compositional semantics of NP, this explained the constraint on determiners.

Beyond these points, we covered a good deal of empirical ground, and sketched analyses of several additional phenomena. Here we find the lexical issues discussed in section 9 particularly interesting. The next step in the investigation which we have begun here is to relate these ideas to the semantics proposed in sections 5 and 6.

Although we outlined a compositional analysis, we did not state it in any concrete way. This was partially because we wished to concentrate on exploring data, but also for technical reasons: in hunting through the literature on dynamic systems for natural language interpretation, we did not find a tool kit which included everything we needed (file discourse referents, presupposition, typed compositional interpretation, possible worlds, and a model theory for files allowing for files in their domains). For the sub-clausal compositional problem discussed in section 8, we find Muskens' system attractive (Muskens 1996), but this is an extensional semantics which does not provide file discourse referents. The theory of files, file updates, and file anaphora we need is provided in Guerts (1995) and Frank (1996). Beaver (1995) provides a typed system of interpretation covering presupposition; another typed compositional treatment of presupposition is given in chapter 4 of Chierchia (1995). Hopefully it will prove possible to put together the package we need from these pieces.

**Endnotes**

*We would like to thank Ede Zimmermann for useful comments on a draft of this paper.

[1] Groenendijk, Stokhof, and Veltman's model theory is based on referent systems (Vermeulen 1994). The assignment mapping variables to individuals is replaced by a function r mapping variables to pegs and a function g mapping pegs to individuals. This results in a dynamic semantics with a (not merely propositional) notion of extension of information, such that updates with formulas are informationally monotonic.

[2] This way of proceeding is not totally divorced from reality, since there are circumstances where a model theory based on such assumptions does not go astray. For instance, suppose we want to model information about the situation in our institute at a given moment (certain people being in the seminar room, certain computers being up or down, the heating being on or off), and the information that different people and computers have about such questions. Then stipulating identification of people and computers across possible situations can be justified on the grounds of having no impact on the logical and semantic questions we are interested in at the moment. For general purposes, though, we prefer an ontology where individuals are not shared across possible worlds.

[3] This requires that a discourse referent be totally identified in order to count as identified, and counts a discourse referent as unidentified if it is partially but not totally identified. Consider a rumour that Solange is having an affair with an Italian, either Sandro or Antonio. Does the sentence 'Solange is having an affair with an unidentified Italian' truthfully describe the rumour? Our analysis says it does. We would have to say that any oddness is due to a quantity implicature.

[4] In the file incrementation term p+f, we allow p to be either a proposition or a file. Alternatively, one can use a construction which identifies propositions with files having empty domains, as in Dekker (1993).

[5] Notice that conditions p:content($x_4$) + f, content($x_4$) ≈ p, Ud($x_5$,p)
are nearly equivalent to the single condition content($x_4$) ≈ content($x_4$) + f + Ud($x_5$)
where Ud is the one-place operator defined in (22). The reason this is so is that Ud($x_5$) functions as a test. If the dref $x_5$ is identified in content($x_4$) + f, the test fails, and content($x_4$) + f + Ud($x_5$) is the contradictory file. The world projection of this is the empty set, meaning that the equivalence fails (as long as content($x_4$) is not empty). This suggests that it might be possible to combine the attitude description with the undefinedness predication. However, there are examples where the file argument cannot be taken to be the output file for the host clause. In the example below, the underlined phase is understood as pertaining to all releases of information up to the present, not just the July press release. This seem to require some kind of quantification over reports sharing a discourse referent with the July press release.

(i)    In July, there was a brief press release about Solange.
       She had signed with a *still unidentified* Hollywood agent.

[6]The data in the middle column are replicated in examples where *the press release* is subject:
   (i)   The press release discloses which team Solange has agreed to play for.
   (ii)  ??The press release determines which team Solange has agreed to play for.
   (iii) The press release specifies which team Solange has agreed to play for.

[7]In the analysis of Abusch, (1993), the definite description could get scope by a combination of existential closure and storage of a restricting property, rather than structural scoping. This would not affect the present discussion in any substantial way.

[8]There is a difference between (42) and (43) relating to where the formula *team-in-italy($x_1$)* enters the interpretation. (42) but not (43) has the agreement carrying the information that the played-for entity is an Italian team.

[9]We note in passing that in these examples the pronoun subject refers to an individual (or, in our analysis, an individual-level discourse referent). The pronoun *this* (and perhaps marginally *it*) can refer to a proposition, but gives us a different reading which includes the existential quantifier:

( i)  Solange hid the money in a closet. This is *unknown*.

This means that it is unknown that she hid the money in a closet, rather than that it is unknown that she hid the money there; it is the latter which expresses the reading which is under discussion in this paper.

[10]Another pattern of data which is relevant here is the paradigm of clausal complementation (see Ginzburg (1996) for a survey). Verbs such as *disclose* and *know* are resolutive/factive verbs; they take both WH and *that* complements:
(i)    I know who ate the cookies. I know that Scott ate the cookies.
(ii)   Spy did not disclose who was involved in the bribery scheme.
       Spy disclosed that a European film star was involved in the bribery scheme.

Verbs in the *name/identify* class somewhat marginally take WH complements, but do not take *that* complements:

(iii)     a.    I won't name who they are.
            b. *I named that they were John, Dick and Harry.

(iv)     a.    I listed who they were.
            b. *I listed that they were John, Dick and Harry.

(v)     a.    I enumerated who they were.
            b. *I enumerated that they were John, Dick and Harry.

(vi)     a.    ?I identified who sneaked in without a ticket.
            b. *I identified that John sneaked in without a ticket.

[11]The status of *another* itself is a whole nother problem.

IMS
Universität Stuttgart
Azenbergstr. 12
70174 Stuttgart
Germany
dorit@ims.uni-stuttgart.de
mats@ims.uni-stuttgart.de